\documentclass[11pt]{article}
\usepackage[preprint]{acl}
\usepackage{amsmath}
\usepackage{amssymb} 
\usepackage{newtxtext, newtxmath}
\usepackage{latexsym}
\usepackage[T1]{fontenc}
\usepackage[utf8]{inputenc}
\usepackage{microtype}
\usepackage{inconsolata}
\usepackage{graphicx}
\usepackage{enumitem}  
\usepackage[breakable]{tcolorbox}
\usepackage{booktabs}
\usepackage{multirow}
\usepackage{algorithm}
\usepackage{algorithmic}
\usepackage{subcaption}
\usepackage{xcolor}
\usepackage{hyperref}

\newcommand{\modelname}{VeriGUI}

\title{%
  \raisebox{-0.3em}{\includegraphics[height=1.5em]{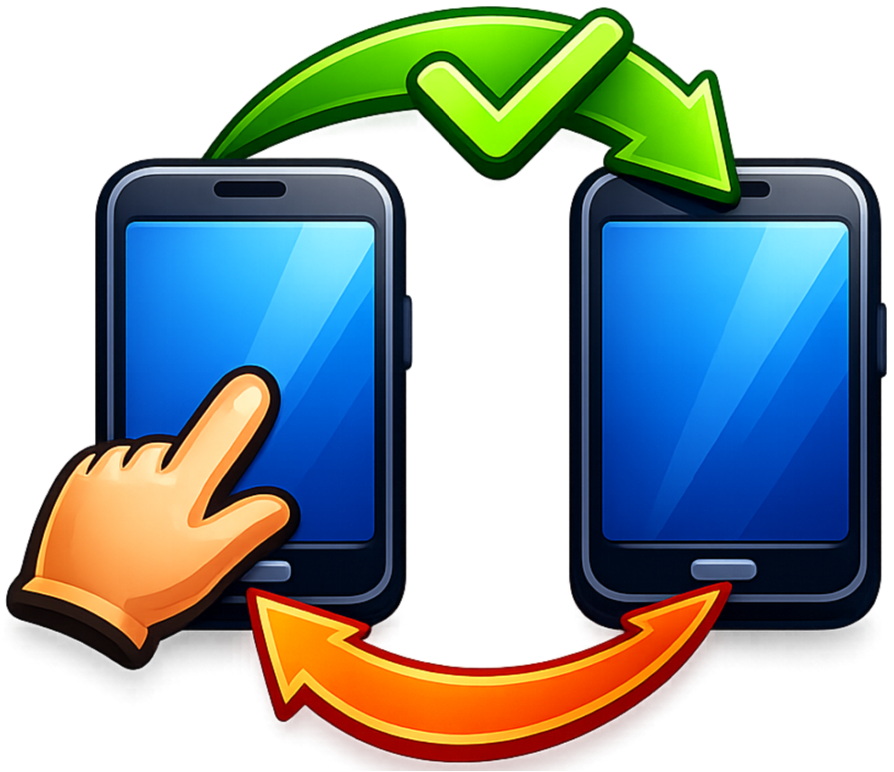}}%
  ~ Don't Act Blindly: Robust GUI Automation via Action-Effect \\
  Verification and Self-Correction%
}

\author{
  \textbf{Yuzhe Zhang}$^{1,2}$\thanks{Work done during the internship at Baidu.} \quad
  \textbf{Xianwei Xue}$^{2}$ \quad
  \textbf{Xingyong Wu}$^{2}$ \quad
  \textbf{Mengke Chen}$^{2}$ \quad
  \textbf{Chen Liu}$^{2}$ \\
  \textbf{Xinran He}$^{2}$ \quad
  \textbf{Run Shao}$^{2}$ \quad
  \textbf{Feiran Liu}$^{1}$ \quad
  \textbf{Huanmin Xu}$^{2}$ \quad
  \textbf{Qiutong Pan}$^{2}$ \quad
  \textbf{Haiwei Wang}$^{2}$\thanks{Corresponding author.} \\[0.5em]
  $^{1}$Beijing University of Technology, Beijing, China \quad
  $^{2}$Baidu Inc., Beijing, China \\[0.3em]
  \texttt{\{zhangyuzhe02, xuexianwei, wuxingyong, chenmengke, liuchen26\}@baidu.com} \\
  \texttt{\{hexinran, shaorun, xuhuanmin, panqiutong, wanghaiwei\}@baidu.com}
}

\begin{document}
\maketitle
\begin{abstract}
Autonomous GUI Agents based on vision-language models (VLMs) often assume deterministic environment responses, generating actions without verifying whether previous operations succeeded. In real-world settings with network latency, rendering delays, and system interruptions, this assumption leads to undetected action failures, repetitive ineffective behaviors, and catastrophic error accumulation. Moreover, learning robust recovery strategies is challenging due to the high cost of online interaction and the lack of real-time feedback in offline datasets. We propose \modelname{} (\textbf{Veri}fication-driven \textbf{GUI} Agent), which explicitly models action outcomes and recovery under noisy environments. \modelname{} introduces a \textbf{T}hinking--\textbf{V}erification--\textbf{A}ction--\textbf{E}xpectation (\textbf{TVAE}) framework to detect failures and guide corrective reasoning, and a two-stage training pipeline that combines \textbf{Robust SFT} with synthetic failure trajectories and \textbf{GRPO} with asymmetric verification rewards. We train \modelname{} in both 3B and 7B configurations. We further construct a \textbf{Robustness Benchmark} based on AndroidControl-High to evaluate failure recognition and correction. Experiments demonstrate that \modelname{}-3B and -7B achieve strong results across offline and online benchmarks, with the verification-and-recovery mechanism transferring effectively to dynamic real-world environments.
\end{abstract}
\section{Introduction}
\label{sec:intro}

\begin{figure}[t]
    \centering
    \vspace{1.5em} 
    \includegraphics[width=0.85\columnwidth]{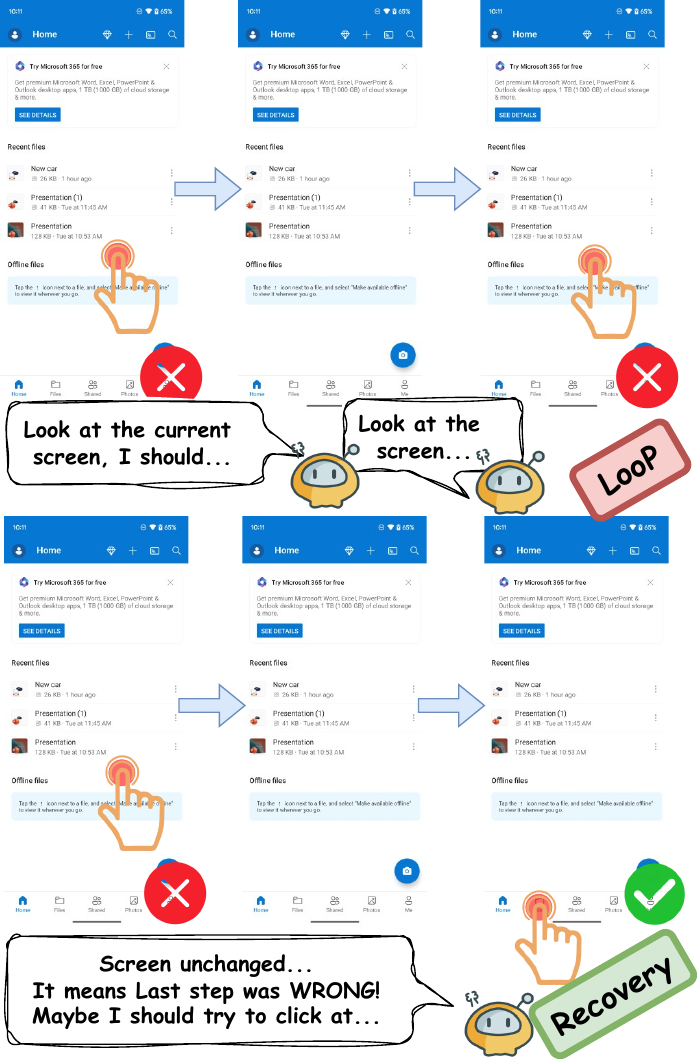}
    \caption{Overview of \modelname{}'s closed-loop framework. Unlike blind action agents (above) that repeat failed actions indefinitely, \modelname{} (below) verifies each action's outcome against predicted effects, enabling failure detection and reasoned recovery.}
    \label{fig:overview}
\end{figure}
GUI automation has emerged as a promising application of vision-language models (VLMs)~\citep{hong2024cogagent, cheng2024seeclick, lin2024showui}, enabling autonomous agents to interpret screenshots, understand natural language instructions, and execute complex multi-step tasks on mobile devices~\citep{zhang2025appagent, rawles2023androidwildlargescaledataset}. These capabilities have driven rapid progress in benchmark performance~\citep{zhou2024webarena, xie2024osworld, rawles2025androidworld} and broadened the scope of automated interaction. However, deploying such agents in real-world environments exposes a fundamental weakness shared by most existing approaches: they implicitly assume that every issued action executes as intended.

This assumption rarely holds in practice. Network latency may prevent a page from loading; rendering delays can cause click targets to shift; system interruptions may block expected transitions or interrupt animations. When such failures occur, current agents continue operating as if nothing went wrong—they observe an unchanged screen yet generate another action based on flawed assumptions about their progress~\citep{huang2024large}. Worse still, because these agents have rarely encountered failure scenarios during training, they tend to repeat the exact same ineffective action, creating infinite execution loops that waste computational resources without advancing the task. Importantly, this class of idempotent failures---where the erroneous action leaves the screen unchanged---is not a corner case. Empirical evidence confirms their prevalence: execution timeouts caused by repeated ineffective actions account for 72.3\% of all failures across 1,265 task executions~\citep{liu2026memguibenchbenchmarkingmemorymobile}; the Reasonable Operation Ratio metric was introduced in AndroidLab precisely to capture such non-progressive operations~\citep{xu-etal-2025-androidlab}; and failed grounding and stuck repetitive behaviors have been identified as primary failure modes in real mobile deployments~\citep{li2025mobileuseguiagenthierarchical}.

Human users naturally avoid this trap. After each interaction, we implicitly verify whether the expected change occurred: did the button highlight, did the page navigate, or did new content appear? If something seems wrong, we pause, diagnose the issue, and adjust our approach accordingly. This verification–reasoning–correction loop forms the basis of robust human-computer interaction~\citep{wei2022chain, yao2023react}, yet it is conspicuously absent from current GUI Agent designs.

Training agents to exhibit such behavior presents its own challenges. Online reinforcement learning in live GUI environments suffers from prohibitive interaction latency, system instability, and poor scalability, making large-scale training impractical---requiring up to 64 parallel Android emulators~\citep{bai2024digirl} or complex distributed infrastructure~\citep{wang2025distrl}. Offline datasets enable faster iteration but lack explicit feedback signals indicating whether an action failed and how recovery should proceed~\citep{li2024effects, rawles2023androidwildlargescaledataset}. This gap between training conditions and deployment realities fundamentally limits the robustness of existing approaches. Prior work on self-correction in LLMs~\citep{madaan2023selfrefine, shinn2023reflexion, gou2024critic} has shown promise in iterative refinement through verbal feedback, but these methods typically assume access to external feedback signals or oracle information, and have not been effectively adapted to the visual grounding requirements of GUI automation~\citep{kumar2025score}.

We propose \modelname{} (\textbf{Veri}fication-driven \textbf{GUI} Agent) to address these limitations. Our approach makes three contributions:
\begin{itemize}
    \item We propose a T-V-A-E framework that forms a closed reasoning loop, enabling outcome verification, failure diagnosis, and expected effect prediction to prevent blind error accumulation.

    \item We design a \textbf{two-stage training pipeline} combining Robust SFT on synthetic failure trajectories with GRPO that leverages GUI failure idempotency to simulate online feedback from offline data, using asymmetric penalties to promote honest self-assessment. This approach avoids the prohibitive infrastructure demands of online RL while still producing effective self-correcting behavior.

    \item We construct a \textbf{Robustness Benchmark} with controlled failure-injection cases and two novel metrics (Loop Rate and Recovery Success Rate) to reveal failure modes hidden in conventional success-only evaluations. On this benchmark, \modelname{}-3B achieves a Recovery Success Rate of 51.1\%, and \modelname{}-7B achieves 52.5\%, both surpassing all tested open-source baselines.
\end{itemize}
\section{Related Work}
\label{sec:related}

\subsection{Vision-Language Models for GUI Agents}

Early GUI Agents relied on structured interface representations such as HTML trees or accessibility metadata, limiting robustness across heterogeneous platforms \citep{deng2024mind2web, zhang2025appagent, zheng2024gpt4vision}. Recent VLMs enable direct visual grounding from raw screenshots, substantially improving cross-platform generalization: CogAgent \citep{hong2024cogagent} and SeeClick \citep{cheng2024seeclick} demonstrate accurate UI element localization and action prediction without structured inputs , while ShowUI \citep{lin2024showui}, Ferret-UI \citep{you2024ferret}, and UI-TARS \citep{qin2025uitars}  further enhance efficiency and generalization through UI-specific pretraining and unified vision--language--action modeling. Subsequent work extends to multi-step reasoning and planning via search, reflection, and execution-feedback-driven replanning \citep{yu2025exact, antoniades2024swesearch, zhang2025webpilot, wu2025reachagent, putta2024agentqadvancedreasoning, wang2025mobileagenteselfevolvingmobileassistant, wu2025backtrackagentenhancingguiagent}, but typically assumes that selected actions execute as intended at the step level and handles failures implicitly through replanning. In contrast, \modelname{} explicitly models action--effect consistency by verifying whether each action induces its intended visual outcome, enabling principled step-level failure detection and recovery.

\begin{figure*}[t]
    \centering
    \includegraphics[width=\textwidth]{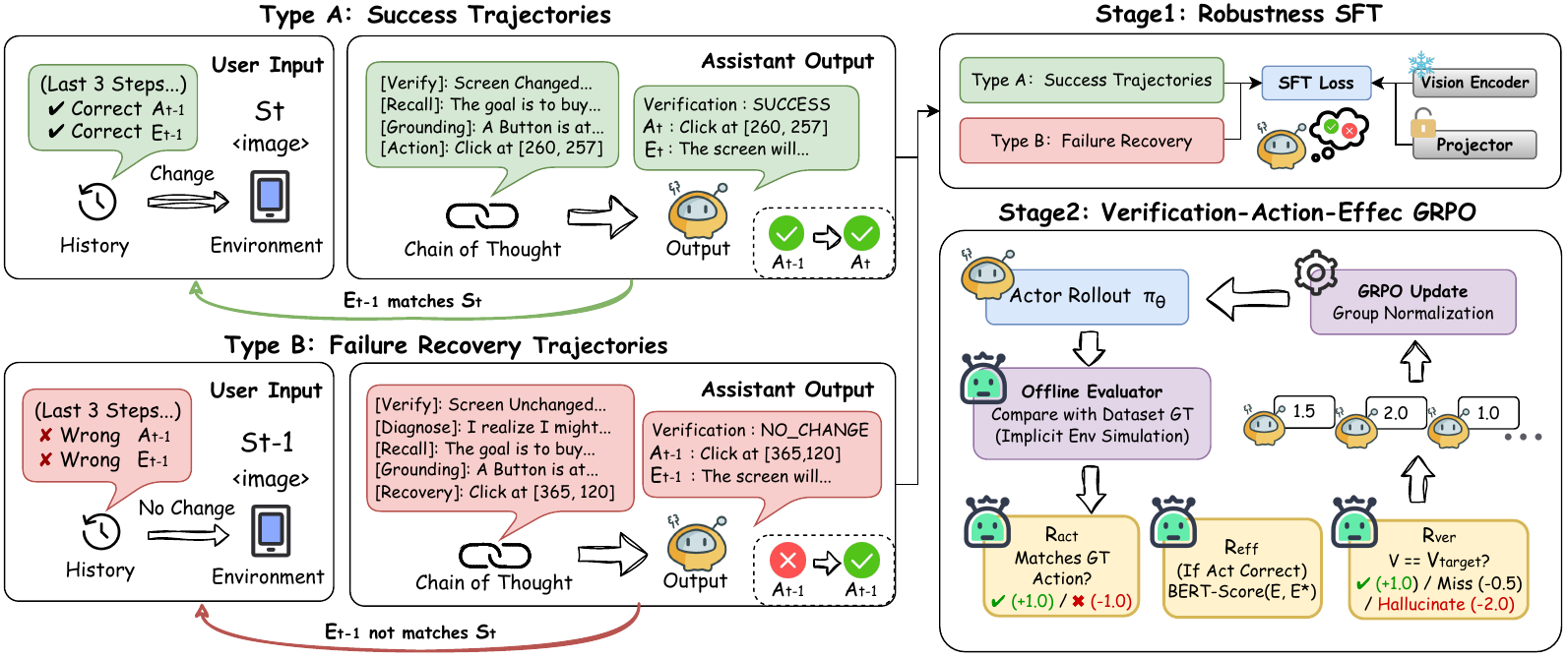}
    \caption{Overview of \modelname{}'s architecture and training pipeline. \textbf{(A)} The TVAE inference cycle implements a closed-loop verification mechanism where structured thinking precedes verification. \textbf{(B)} The two-stage training pipeline: Stage 1 establishes basic verification capabilities via Robust SFT on mixed success/failure trajectories; Stage 2 refines self-correction via GRPO, using offline data to simulate online feedback through composite rewards.}
    \label{fig:architecture}
\end{figure*}

\subsection{Reinforcement Fine-Tuning and Learning for GUI Agents}

Reinforcement learning (RL) has become central to improving GUI Agent robustness and generalization, evolving from trajectory-level optimization in DigiRL \citep{bai2024digirl} and DistRL \citep{wang2025distrl} to preference-based learning with Direct Preference Optimization (DPO) \citep{rafailov2024direct} and self-generated supervision \citep{kumar2025score, welleck2023generating}. For long-horizon reasoning, Group Relative Policy Optimization (GRPO) provides a stable and scalable framework underlying recent reasoning-oriented models \citep{deepseek2025r1}. Building on this line of work, GRPO has been adopted in GUI Agents such as UI-R1 \citep{lu2025uir1}, which applies it to unified action spaces, and InfiGUI-R1 \citep{liu2025infiguiagent}, which integrates reactive execution with deliberative reasoning . However, most RL-based GUI Agents optimize task-level success or proxy step correctness, while \modelname{} treats action--effect verification as a first-class reinforcement objective, enabling the agent to learn honest self-monitoring and robust self-correction.
\section{Methodology}
\label{sec:method}

We present \modelname{}, a Verification-driven GUI Agent that addresses the fundamental limitation of current GUI automation systems: their inability to verify execution outcomes and recover from failures. As illustrated in Figure~\ref{fig:architecture}, our approach combines a closed-loop inference cycle with a two-stage training pipeline that systematically develops verification-driven self-correction capabilities.

\subsection{Problem Formulation}
\label{sec:problem}

We formalize GUI automation as a sequential decision-making process where an agent must execute interactions to satisfy a natural language instruction $I$. At each timestep $t$, the agent observes a screen state $S_t$ and an action history $H_t$, and selects an action $A_t$ from a discrete action space $\mathcal{A} = \{\texttt{click}, \texttt{scroll}, \texttt{input\_text}, \texttt{long\_press}...\}$ according to policy $\pi_\theta(A_t \mid S_t, H_t, I)$.

Unlike conventional approaches that assume deterministic transitions, we explicitly model execution uncertainty. We augment the policy to jointly predict structured Thinking $T_t$, verification judgment $V_t$, action $A_t$, and expected effect $E_t$. This formulation enables the agent to detect failures by verifying $V_t$ against visual evidence and self-correct without external supervision.

\subsection{TVAE Inference Cycle}
\label{sec:tvae}

At the core of \modelname{} lies a structured reasoning process mirroring human interaction patterns (Figure~\ref{fig:architecture}A). Crucially, TVAE is not a linear chain but a temporally linked cycle: the expected effect predicted at step $t$ becomes the verification hypothesis at step $t{+}1$. This temporal dependency enforces causal consistency across steps, ensuring that errors cannot be ignored or overwritten by subsequent actions. At each step $t$, the agent produces:

\paragraph{1. Think ($T_t$).} Structured analysis using explicit tags (\texttt{[Verify]}, \texttt{[Recall]}, \texttt{[Grounding]}, \texttt{[Action]}). When correcting errors, the structure shifts to \texttt{[Diagnose]} and \texttt{[Recovery]}, ensuring verification is grounded in visual reasoning. Note that the \texttt{[Verify]} tag within \texttt{<think>} constitutes the internal reasoning process, while the \texttt{<verification>} output below represents the final binary judgment derived from that reasoning.

\paragraph{2. Verification ($V_t$).} A binary judgment assessing the previous step:
\begin{equation}
V_t = \begin{cases}
\text{SUCCESS} & \text{if } S_t \text{ matches } E_{t-1} \\
\text{NO\_CHANGE} & \text{if } S_t \text{ not match } E_{t-1}
\end{cases}
\end{equation}

\paragraph{3. Action ($A_t$).} Executable JSON (for instance, \texttt{\{"action": "click", "coordinate": [x,y]\}}).

\paragraph{4. Expected Effect ($E_t$).} A prediction of the resulting screen change, serving as the verification target for step $t+1$.

\subsection{Stage 1: Robust Supervised Fine-Tuning}
\label{sec:sft}

To teach error recognition, we construct a mixed dataset of positive and synthetic negative samples:

\textbf{Type A: Success Trajectories.} Standard steps where $S_t$ reflects the successful execution of $A_{t-1}$. The target output confirms success (\texttt{SUCCESS}) and proceeds with the next ground-truth action.

\textbf{Type B: Failure Recovery Trajectories.} We simulate failures by pairing the previous screen $S_{t-1}$ with history claiming action $A_{t-1}$ was executed. This creates a "no change" scenario. The target output requires the agent to diagnose the failure (\texttt{NO\_CHANGE}) and generate a corrective action (typically a refined $A_{t-1}^*$).

We employ GPT-4o to generate structured Chain-of-Thought (CoT) annotations for both types and train using standard cross-entropy loss(Appendix~\ref{app:sft_data}). This stage serves as a necessary behavioral prior: without explicit exposure to failure cases during imitation learning, the model tends to overfit to optimistic assumptions that all actions succeed, making subsequent reinforcement learning unstable.

\subsection{Stage 2: Verification-Action-Effect GRPO}
\label{sec:grpo}

While SFT establishes basic capabilities, it lacks the feedback loop required to internalize "honest" verification. Stage 2 employs Group Relative Policy Optimization (GRPO) with a specialized reward mechanism that simulates online feedback using only offline data.

\paragraph{Implicit Environment Simulation.}
Direct reinforcement learning in live GUI environments suffers from prohibitive interaction latency, system instability, and scalability bottlenecks---requiring up to 64 parallel Android emulators~\citep{bai2024digirl} or complex distributed infrastructure~\citep{wang2025distrl}. To bypass these inefficiencies while preserving realistic feedback dynamics, we employ a \textbf{data-driven implicit simulation strategy}. We leverage the idempotency property of GUI errors: \textit{incorrect actions typically leave the screen unchanged}. This assumption is well-grounded in practice---execution timeouts from repeated ineffective actions, screen-invariant operation rates, and failed grounding with no state transition together constitute the dominant failure modes in real mobile environments~\citep{xu-etal-2025-androidlab, liu2026memguibenchbenchmarkingmemorymobile, li2025mobileuseguiagenthierarchical}---and enables us to transform offline trajectories into training samples with meaningful environment feedback signals without any live interaction.

We construct the RL training data from the mixed SFT dataset. For any given input state (which may represent a successful or failed history), the model generates a response. We then determine the verification ground truth $V_{\text{target}}$ dynamically based on the sample type:
\begin{itemize}
    \item For Type A inputs (Success History), $V_{\text{target}} = \text{SUCCESS}$.
    \item For Type B inputs (Failure History), $V_{\text{target}} = \text{NO\_CHANGE}$.
\end{itemize}
Crucially, the model is rewarded only if its predicted verification $\hat{V}_t$ matches this objective reality, regardless of whether it "thinks" it succeeded. This effectively simulates environment feedback: if the screen didn't change (Type B), the environment "tells" the agent it failed by rewarding only \texttt{NO\_CHANGE} predictions. This implicit feedback mechanism is intentionally minimalistic: it does not reveal the correct action or recovery strategy, but only signals whether the agent's belief about the outcome is consistent with visual reality. As a result, the agent must rely on its own reasoning and expected-effect predictions to determine how to proceed, closely mirroring real-world deployment conditions.

\paragraph{Composite Reward Function.}

The reward function $R_t$ drives the optimization of the TVAE cycle components:
\begin{equation}
\label{eq:reward}
R_t = R_{\text{act}} + \alpha \cdot R_{\text{eff}} + \beta \cdot R_{\text{ver}}
\end{equation}

Together, these reward components explicitly couple action execution with outcome awareness, ensuring that high task completion scores cannot be achieved by systematically ignoring verification signals.

1. \textbf{Action Reward ($R_{\text{act}}$):} Measures execution correctness by comparing $\hat{A}_t$ against the ground-truth action $A_t^*$ (using type matching and coordinate IoU).
\begin{equation}
    R_{\text{act}} = \mathbb{I}(d(\hat{A}_t, A_t^*) \leq \delta) - \mathbb{I}(d(\hat{A}_t, A_t^*) > \delta)
\end{equation}

2. \textbf{Effect Reward ($R_{\text{eff}}$):} Evaluates semantic consistency between the predicted expected effect $\hat{E}_t$ and the reference effect $E_t^*$ generated by GPT-4o during dataset construction (Appendix~\ref{app:sft_data}). $R_{\text{eff}}$ is computed via BERTScore when the action is correct ($d(\hat{A}_t, A_t^*) \leq \delta$), and set to $0$ otherwise:
\begin{equation}
    R_{\text{eff}} = \begin{cases}
    \text{BERTScore}(\hat{E}_t,\, E_t^*) & \text{if } d(\hat{A}_t, A_t^*) \leq \delta \\
    0 & \text{otherwise}
    \end{cases}
\end{equation}

3. \textbf{Verification Reward ($R_{\text{ver}}$):} Enforces honest self-monitoring with asymmetric penalties:
\begin{equation}
    R_{\text{ver}} = \begin{cases}
   +1.0 & \text{if } \hat{V}_t = V_{\text{target}} \\
   -0.5 & \text{if False Negative (Miss)} \\
   -2.0 & \text{if False Positive (Hallucination)}
   \end{cases}
\end{equation}
   The severe penalty for hallucination ($-2.0$) forces the agent to align its internal belief with visual reality, preventing the accumulation of errors.

\paragraph{Optimization.}
We employ Group Relative Policy Optimization (GRPO)~\citep{shao2024deepseekmath} to optimize the policy. For each input $x$, GRPO samples a group of $G$ outputs $\{y_1, \dots, y_G\}$ from the current policy $\pi_\theta$. Let $R_t(y_i)$ denote the composite reward defined in Eq.~\ref{eq:reward} evaluated for output $y_i$. The group-normalized advantage for the $i$-th output is:
\begin{equation}
\hat{A}_{i} = \frac{R_t(y_i) - \mathrm{mean}\bigl(\{R_t(y_j)\}_{j=1}^{G}\bigr)}{\mathrm{std}\bigl(\{R_t(y_j)\}_{j=1}^{G}\bigr) + \varepsilon}
\end{equation}
where $\varepsilon$ is a small constant for numerical stability. For each token position $k$ in output $y_i$, we define the probability ratio between the updated policy and the old sampling policy as:
\begin{equation}
\rho_{i,k} = \frac{\pi_\theta(y_{i,k} \mid y_{i,<k},\, x)}{\pi_{\theta_{\text{old}}}(y_{i,k} \mid y_{i,<k},\, x)}
\end{equation}
The clipped surrogate loss for each token is then:
\begin{equation}
\mathcal{L}^{\text{clip}}_{i,k} = \min\!\left(\rho_{i,k}\,\hat{A}_{i},\ \mathrm{clip}(\rho_{i,k},\,1{-}\epsilon_{\text{clip}},\,1{+}\epsilon_{\text{clip}})\,\hat{A}_{i}\right)
\end{equation}
where $\epsilon_{\text{clip}}$ restricts the probability ratio to $[1{-}\epsilon_{\text{clip}},\, 1{+}\epsilon_{\text{clip}}]$. The final GRPO objective aggregates the clipped surrogate loss over all outputs and tokens, with a KL penalty constraining deviation from the reference policy $\pi_{\text{ref}}$:
\begin{equation}
\mathcal{J}_{\text{GRPO}} = \mathbb{E}_{x}\!\left[\frac{1}{G}\sum_{i=1}^{G}\frac{1}{L_i}\sum_{k=1}^{L_i} \mathcal{L}^{\text{clip}}_{i,k}\right] - \lambda\, D_{\text{KL}}(\pi_\theta \| \pi_{\text{ref}})
\end{equation}
where $L_i$ is the token length of the $i$-th output and $\lambda$ controls the KL regularization strength.
The complete inference and training procedures are detailed in Algorithms~\ref{alg:tvae_inference} and \ref{alg:training} (Appendix~\ref{app:algorithm}).
\section{Experiments}
\label{sec:experiments}

We evaluate \modelname{} across multiple benchmarks: AndroidControl-High \citep{li2024effects}, AITW-Gen \citep{rawles2023androidwildlargescaledataset}, and GUI Odyssey \citep{lu2025guiodysseycomprehensivedatasetcrossapp} as our offline evaluation suite, and MiniWoB++ \citep{liu2018reinforcementlearningwebinterfaces} and AndroidWorld \citep{rawles2025androidworld} as fully online environments to validate real-world robustness.

\subsection{Experimental Setup}

\paragraph{Dataset and Baselines.}
AndroidControl-High provides approximately 15,000 training trajectories and 2,500 test trajectories spanning productivity, social media, e-commerce, and system applications. We compare against 3B models (Qwen2.5-VL-3B \citep{bai2025qwen25vl}, UI-R1-3B \citep{lu2025uir1}) , 7B/8B models (OS-Genesis-7B \citep{sun-etal-2025-os}, OS-Atlas-7B \citep{wu2024osatlas}, Qwen2.5-VL-7B, AgentCPM-GUI-8B \citep{agentcpm}, UI-TARS-7B \citep{qin2025uitars}, UI-S1-7B \citep{lu2025uis1advancingguiautomation}), and closed-source systems GPT-5.1 and Gemini-3-flash.

\paragraph{Implementation.}
We apply our two-stage training pipeline at two scales: \modelname{}-3B builds on Qwen2.5-VL-3B, and \modelname{}-7B builds on Qwen2.5-VL-7B. Both variants use the same pipeline configuration. Stage 1 trains for 2 epochs with learning rate $1\times10^{-5}$; Stage 2 runs 15 epochs with learning rate $5\times10^{-6}$, group size $G=6$, and reward weights $\alpha=0.5$, $\beta=0.5$. Training uses 8$\times$A100 GPUs. Full hyperparameters appear in Appendix~\ref{app:implementation}.

\subsection{Evaluation Protocol}
\label{sec:eval_protocol}

Our offline evaluation exploits the \textit{failure idempotency} of GUI environments to construct a pseudo-online simulation: correct actions advance to ground-truth next screens, while incorrect actions return the unchanged screen. Formal metric definitions appear in Table~\ref{tab:metrics-summary} in Appendix~\ref{app:metrics}.

\paragraph{Step Level Metrics.}
Step-level metrics assess single-step predictions independently: Type Match (TM), Grounding Rate (GR), and Step Success Rate (SR).

\paragraph{Task Level Metrics.}
Task-level metrics evaluate end-to-end execution under pseudo-online simulation with a $2T_{\text{GT}}$ step budget: Task Success Rate (TSR), Progress (PG), Simulated Task Success Rate (Sim-TSR), and Average Step Overhead (ASO).

\paragraph{Robustness Metrics.}
We construct failure-injection test cases by pairing unchanged screens with histories claiming an action was executed. Loop Rate (LR) measures repeated failed actions; Recovery Success Rate (RSR) measures correct recovery actions.

\subsection{Main Results}

\paragraph{Offline Benchmark.}
Table~\ref{tab:main-results} presents comprehensive results across all offline benchmarks. On AndroidControl-High, \modelname{}-3B achieves Type Match of 72.2\%, the highest among all open-source models at the 3B scale, and \modelname{}-7B reaches 74.2\%, establishing a new open-source state of the art. Notably, \modelname{}-3B surpasses several 7B-scale baselines on this metric despite having less than half the parameters, which demonstrates that the TVAE framework's structured reasoning actively improves action type selection rather than merely adding overhead.

The task-level metrics reveal the practical value of verification-driven self-correction most clearly. All 3B-scale baselines without explicit verification fail entirely under pseudo-online conditions, achieving zero Sim-TSR, while larger 7B-scale models achieve modest results. \modelname{}-3B completes 16.7\% of tasks with an ASO of just 1.25, and \modelname{}-7B further improves to 23.5\% Sim-TSR with an ASO of 1.09---both substantially outperforming all comparably sized open-source models. The gap between TSR and Sim-TSR directly quantifies tasks completed through error recovery rather than flawless execution.

Cross-distribution results on AITW-Gen and GUI Odyssey further validate generalization. \modelname{}-7B achieves 72.7 TM and 52.3 SR on GUI Odyssey, exceeding UI-TARS-7B on both metrics and closely approaching UI-S1-7B's SR of 52.8 while surpassing it on TM. \modelname{}-3B shows consistent improvements over same-scale baselines on GUI Odyssey across all metrics, and achieves higher Sim-TSR than both 3B baselines on AITW-Gen, confirming that the TVAE framework provides transferable benefits.

\begin{table*}[t]
\centering
\small
\resizebox{\textwidth}{!}{%
\begin{tabular}{llccccccc|cc|ccc}
\toprule
\multirow{2}{*}{\textbf{Category}} & \multirow{2}{*}{\textbf{Model}}
  & \multicolumn{7}{c|}{\textbf{AndroidControl-High}}
  & \multicolumn{2}{c|}{\textbf{AITW-Gen}}
  & \multicolumn{3}{c}{\textbf{GUI Odyssey}} \\
\cmidrule(lr){3-9} \cmidrule(lr){10-11} \cmidrule(lr){12-14}
\cmidrule(lr){3-5} \cmidrule(lr){6-9}
& & TM & GR & SR & TSR & PG & Sim-TSR & ASO$\downarrow$ & PG & Sim-TSR & TM & GR & SR \\
\midrule
\multirow{2}{*}{\shortstack[l]{Closed-\\source}}
& GPT-5.1       & 70.1 & 30.0 & 23.1 & -- & -- & -- & -- & 10.4 & 2.5 & 68.1 & 34.0 & 27.9 \\
& Gemini-3-flash & 83.3 & 36.9 & 42.2 & -- & -- & -- & -- & 17.8 & 4.5 & 83.2 & 48.8 & 46.9 \\
\midrule
\multirow{2}{*}{\shortstack[l]{Open-src\\3B}}
& Qwen2.5-VL-3B & 60.1 & 52.8 & 31.2 & 0 & 2.8 & 0 & $\infty$ & 1.5 & 0.1 & 62.0 & 47.4 & 39.9 \\
& UI-R1-3B      & 62.4 & 50.4 & 30.3 & 0 & 4.9 & 0 & $\infty$ & 8.8 & 2.4 & 62.5 & 52.6 & 32.0 \\
\midrule
\multirow{6}{*}{\shortstack[l]{Open-src\\7B/8B}}
& Qwen2.5-VL-7B    & 68.5 & 61.7 & 42.8 & 6.2 & 12.8 & 10.6 & 1.15     & 11.7 & 3.0 & 72.3 & 55.8 & 48.5 \\
& OS-Genesis-7B    & 64.5 & 47.5 & 35.7 & 0   & 6.5  & 0    & $\infty$ & 6.5  & 2.2 & 62.3 & 59.9 & 34.5 \\
& OS-Atlas-7B      & 67.7 & 42.2 & 35.5 & 0   & 5.5  & 0    & $\infty$ & 12.3 & 3.4 & 70.2 & 63.5 & 41.3 \\
& AgentCPM-GUI-8B  & 42.2 & \textbf{66.4} & 34.1 & 0 & 7.1 & 0 & $\infty$ & 12.0 & 3.4 & 70.3 & 50.2 & 47.8 \\
& UI-TARS-7B       & 71.9 & 60.6 & 47.7 & 10.0 & 14.7 & 13.3 & 1.67 & 16.8 & \textbf{5.9} & 72.6 & 62.1 & 52.0 \\
& UI-S1-7B         & 70.1 & 53.6 & 44.5 & 13.3 & 25.4 & 16.7 & 1.33 & \textbf{19.5} & 4.6 & 70.5 & 62.8 & \textbf{52.8} \\
\midrule
\multirow{2}{*}{Ours}
& \modelname{}-3B & 72.2 & 56.4 & 46.6 & 13.3 & 23.3 & 16.7 & 1.25 & 7.4 & 2.6 & 65.5 & 54.5 & 43.4 \\
& \modelname{}-7B & \textbf{74.2} & 65.5 & \textbf{51.1} & \textbf{16.7} & \textbf{30.6} & \textbf{23.5} & \textbf{1.09} & 15.5 & 5.5 & \textbf{72.7} & \textbf{63.2} & 52.3 \\
\bottomrule
\end{tabular}%
}
\caption{Offline benchmark results. \modelname{} is trained at both 3B and 7B scales. \textbf{Bold} indicates best among all open-source models. $\infty$ ASO indicates no tasks completed.}
\label{tab:main-results}
\end{table*}

\paragraph{Robustness Benchmark.}
Table~\ref{tab:robust-results} evaluates error recognition and recovery directly. UI-TARS-7B achieves the lowest Loop Rate at 13.4\% but only 45.5\% RSR---it avoids repeating failed actions but often produces incorrect alternatives. OS-Atlas-7B shows high Loop Rate of 43.3\% with poor RSR of 10.7\%, indicating neither recognition nor correction capability.

\modelname{}-3B achieves RSR of 51.1\%, surpassing UI-TARS-7B by over five points despite having less than half the parameters, and \modelname{}-7B further improves to 52.5\%, setting a new open-source best. This validates our core design: the TVAE framework trains the model not merely to detect failures but to reason about their causes and generate appropriate corrections. The Loop Rate of 24.3\% for \modelname{}-3B represents a meaningful reduction from same-scale baselines like Qwen2.5-VL-3B at 30.0\%, while \modelname{}-7B achieves 15.6\%, approaching the best open-source Loop Rate of UI-TARS-7B.

\begin{table}[t]
\centering
\small
\begin{tabular}{llcc}
\toprule
\textbf{Category} & \textbf{Model} & \textbf{LR}$\downarrow$ & \textbf{RSR}$\uparrow$ \\
\midrule
\multirow{2}{*}{\shortstack[l]{Closed-source}}
& GPT-5.1 & 11.0 & 21.9 \\
& Gemini-3-flash & 15.0 & 37.0 \\
\midrule
\multirow{2}{*}{\shortstack[l]{Open-source\\3B}}
& Qwen2.5-VL-3B & 30.0 & 35.0 \\
& UI-R1-3B & 29.5 & 29.0 \\
\midrule
\multirow{4}{*}{\shortstack[l]{Open-source\\7B/8B}}
& Qwen2.5-VL-7B & 23.5 & 34.1 \\
& OS-Atlas-7B & 43.3 & 10.7 \\
& UI-TARS-7B & \textbf{13.4} & 45.5 \\
& UI-S1-7B & 20.5 & 43.6 \\
\midrule
\multirow{2}{*}{Ours}
& \modelname{}-3B & 24.3 & 51.1 \\
& \modelname{}-7B & 15.6 & \textbf{52.5} \\
\bottomrule
\end{tabular}
\caption{Robustness Benchmark results. \textbf{Bold} indicates best among all open-source models; \modelname{} achieves the highest Recovery Success Rate at both scales.}
\label{tab:robust-results}
\end{table}

\paragraph{Online Benchmark Evaluation.}

\begin{table}[t]
\centering
\small
\begin{tabular}{llcc}
\toprule
\textbf{Category} & \textbf{Model} & \textbf{MiniWoB++} & \textbf{AW} \\
\midrule
\multirow{2}{*}{\shortstack[l]{Open-source\\3B}}
& Qwen2.5-VL-3B & 21.2 & 6.1 \\
& UI-R1-3B & 33.4 & 10.9 \\
\midrule
\multirow{2}{*}{\shortstack[l]{Open-source\\7B}}
& Qwen2.5-VL-7B & 47.0 & 14.9 \\
& UI-S1-7B & 56.6 & 22.7 \\
\midrule
\multirow{2}{*}{Ours}
& \modelname{}-3B & 35.6 & 12.6 \\
& \modelname{}-7B & \textbf{59.7} & \textbf{25.1} \\
\bottomrule
\end{tabular}
\caption{Online benchmark results. \textbf{Bold} indicates best among all open-source models; \modelname{}-7B outperforms all open-source baselines; \modelname{}-3B surpasses same-scale models and approaches 7B-scale performance.}

\label{tab:online-results}
\end{table}

To validate robustness beyond pseudo-online simulation, we evaluate \modelname{} on two fully online benchmarks where environmental uncertainty arises organically: MiniWoB++, which executes real web interactions in a live browser involving dynamic rendering and network variability; and AndroidWorld (AW), deployed on Android emulators via Android Studio, where real application execution introduces system-level dynamics. Neither benchmark involves artificial failure injection. Table~\ref{tab:online-results} summarizes the results.

\modelname{}-3B achieves 35.6\% on MiniWoB++ and 12.6\% on AndroidWorld, surpassing all 3B baselines on both benchmarks and approaching the performance of 7B-scale models. \modelname{}-7B reaches 59.7\% and 25.1\% respectively, outperforming all open-source baselines including UI-S1-7B. These results are especially encouraging because failure modes in live environments---navigation errors, rendering races, application-level exceptions---are structurally more diverse than the idempotency-based failures modeled during training. The gains transferring to non-idempotent settings suggest that our verification-and-recovery training instills a general habit of self-checking rather than a narrow pattern tied to unchanged screens.

\begin{figure*}[t]
    \centering
    \includegraphics[width=0.85\textwidth]{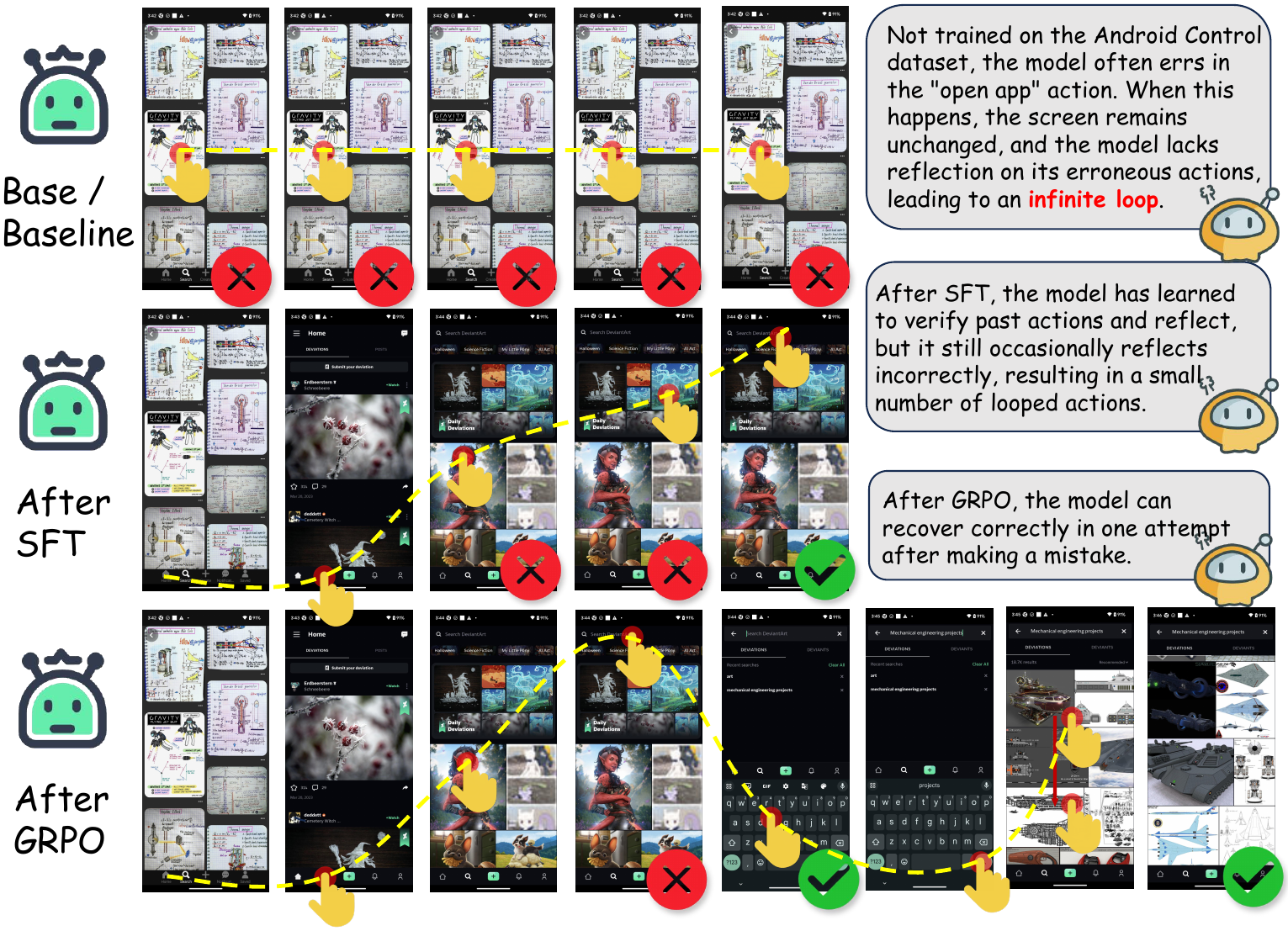}
    \caption{Qualitative comparison across training stages. The base model repeatedly attempts the same ``open app'' action without recognizing failure, entering an infinite loop. After SFT, the model learns to verify and reflect on past actions, though occasional incorrect reflections still cause short loops. After GRPO, the model correctly identifies the failure after one attempt and successfully recovers with an alternative action.}
    \label{fig:case-study}
\end{figure*}

\subsection{Ablation Studies}

The following ablation studies all use \modelname{}-3B on AndroidControl-High to isolate individual design decisions. We use the 3B configuration throughout for computational efficiency and to ensure fair apples-to-apples comparison across variants; the full model's 7B results appear in the main comparison tables above.

\begin{table*}[t]
\centering
\small
\begin{tabular}{lccc|cccc|cc}
\toprule
\multirow{2}{*}{\textbf{Configuration}} & \multicolumn{3}{c|}{\textbf{Step-level (AndroidControl)}} & \multicolumn{4}{c|}{\textbf{Task-level (AndroidControl)}} & \multicolumn{2}{c}{\textbf{Robustness}} \\
\cmidrule(lr){2-4} \cmidrule(lr){5-8} \cmidrule(lr){9-10}
& TM & GR & SR & TSR & PG & Sim-TSR & ASO$\downarrow$ & LR$\downarrow$ & RSR \\
\midrule
Base (Qwen2.5-VL-3B) & 60.1 & 52.8 & 31.2 & 0 & 2.8 & 0 & $\infty$ & 30.0 & 35.0 \\
+ Standard SFT & 67.2 & 56.6 & 42.6 & 6.7 & 16.4 & 10.0 & 3.33 & 30.4 & 29.7 \\
+ Robust SFT & 68.7 & 52.4 & 41.7 & 10.0 & 20.8 & 13.3 & 2.25 & 26.5 & 45.5 \\
+ GRPO (\modelname{}-3B) & \textbf{72.2} & \textbf{56.4} & \textbf{46.6} & \textbf{13.3} & \textbf{23.3} & \textbf{16.7} & \textbf{1.25} & \textbf{24.3} & \textbf{51.1} \\
\bottomrule
\end{tabular}
\caption{Ablation on training stages (\modelname{}-3B, AndroidControl-High). Standard SFT uses only Type A (Success Trajectories) samples, while Robust SFT leverages both Type A and Type B (Failure Recovery Trajectories) samples.}
\label{tab:ablation-stages}
\end{table*}

\paragraph{Training Stage Composition.}
Table~\ref{tab:ablation-stages} isolates each stage's contribution. Standard SFT improves step-level accuracy significantly but provides no verification benefit---Loop Rate actually increases slightly as the model becomes more confident in sometimes incorrect actions. Robust SFT with 30\% synthetic failures introduces genuine error recognition, with RSR jumping from 29.7\% to 45.5\%. GRPO further refines all capabilities; notably ASO drops from 2.25 to 1.25, indicating RL teaches efficient recovery rather than trial-and-error.

\begin{table}[t]
\centering
\small
\begin{tabular}{lcccc}
\toprule
\textbf{Type A:B Ratio} & \textbf{SR} & \textbf{Sim-TSR} & \textbf{LR}$\downarrow$ & \textbf{RSR} \\
\midrule
100:0 (no negatives) & 42.6 & 10.0 & 30.4 & 29.7 \\
90:10 & 42.1 & 10.0 & 29.6 & 30.3 \\
70:30 (default) & 41.7 & 13.3 & 26.5 & \textbf{45.5} \\
50:50 & 39.2 & 12.4 & 27.1 & 45.4 \\
\bottomrule
\end{tabular}
\caption{Effect of negative sample ratio (\modelname{}-3B, AndroidControl-High). The 70:30 balance maximizes RSR while maintaining standard accuracy.}
\label{tab:ablation-ratio}
\end{table}

\paragraph{Negative Sample Ratio.}
Table~\ref{tab:ablation-ratio} examines the sample ratio in Robust SFT. Without failure samples, no verification capability develops. RSR peaks at 45.5\% with the 70:30 ratio; excessive negatives at 50:50 degrade Step Accuracy while yielding diminishing robustness gains.

\begin{table}[t]
\centering
\small
\begin{tabular}{lcccc}
\toprule
\textbf{Reward Components} & \textbf{SR} & \textbf{Sim-TSR} & \textbf{LR}$\downarrow$ & \textbf{RSR} \\
\midrule
$R_{\text{act}}$ only & 42.9 & 13.6 & 25.5 & 44.9 \\
$R_{\text{act}} + R_{\text{ver}}$ & 44.2 & 15.1 & 25.4 & 48.1 \\
$R_{\text{act}} + R_{\text{eff}} + R_{\text{ver}}$ & \textbf{46.6} & \textbf{16.7} & \textbf{24.3} & \textbf{51.1} \\
\bottomrule
\end{tabular}
\caption{Ablation on reward components (\modelname{}-3B, AndroidControl-High). All three components contribute to the final performance.}
\label{tab:ablation-reward}
\end{table}

\paragraph{Reward Components.}
Table~\ref{tab:ablation-reward} analyzes the composite reward. Action reward alone achieves 44.9\% RSR. Adding the verification reward with asymmetric penalties improves RSR to 48.1\%, and the full reward including effect prediction reaches best performance across all metrics, validating our design to treat effect prediction as a first-class TVAE component.

\section{Analysis}
\label{sec:analysis}

\subsection{Case Study}
\label{sec:case_study}

Figure~\ref{fig:case-study} illustrates the progressive improvement in failure handling across training stages through a concrete example. When an ``open app'' action fails due to misaligned coordinates, the base model observes an unchanged screen but lacks any mechanism to recognize this discrepancy. Without verification capability, it repeats identical actions indefinitely, entering the infinite loop pattern we identified as the dominant failure mode in baseline agents.

After Robust SFT, the model acquires basic verification capability through exposure to synthetic failure scenarios. It can sometimes recognize that the screen has not changed and attempt different actions, but occasional incorrect assessments still produce short loops where the model fails to correctly diagnose the root cause. After GRPO, the model reliably detects the unchanged screen, correctly diagnoses that the coordinate error caused the failure, and generates a successful recovery action on the first attempt. The asymmetric penalty structure during RL training teaches the model that overconfident success claims carry severe consequences, while honest uncertainty acknowledgment is acceptable. This progression validates our two-stage design: SFT establishes the structural capability for verification, while GRPO refines its accuracy and reliability.

\subsection{Computational Overhead}
\label{sec:overhead}

The TVAE framework introduces additional output tokens for verification, reasoning, and effect prediction. Table~\ref{tab:efficiency} reports trajectory-level efficiency measured over 50 AndroidControl-High trajectories. Relative to same-scale baselines, per-step output length increases by approximately 45\% for both \modelname{}-3B and \modelname{}-7B. However, this overhead is substantially amortized at the trajectory level because the recovery mechanism breaks failure loops early.

\begin{table}[t]
\centering
\resizebox{\columnwidth}{!}{
\begin{tabular}{lcccc}
\toprule
\textbf{Model} & \textbf{Tok/Step} & \textbf{Tok/Traj} & \textbf{Time/Step (s)} & \textbf{Time/Traj (s)} \\
\midrule
Qwen2.5-VL-3B & 76.3 & 847 & 1.06 & 12.1 \\
Qwen2.5-VL-7B & 81.5 & 724 & 1.29 & 11.3 \\
\modelname{}-3B & 110.8 & 896 & 1.85 & 15.3 \\
\modelname{}-7B & 117.2 & 712 & 2.40 & 14.6 \\
\bottomrule
\end{tabular}
}
\caption{Trajectory-level efficiency on AndroidControl-High (50 trajectories). Per-step token and time costs increase with TVAE, but recovery reduces total trajectory length, keeping trajectory-level overhead modest.}
\label{tab:efficiency}
\end{table}

As shown in Table~\ref{tab:efficiency}, \modelname{}-3B's trajectory-level time rises by only $\sim$26\% relative to Qwen2.5-VL-3B, while Tok/Trajectory increases by just 6\%. More strikingly, \modelname{}-7B's Tok/Trajectory of 712 falls \textit{below} Qwen2.5-VL-3B's 847: stronger recovery eliminates more failure loops than the additional reasoning tokens cost. Furthermore, all model inference in our deployment architecture runs cloud-side; the mobile client handles only rendering and execution, so additional reasoning tokens do not directly affect device battery or latency. Overall, \modelname{} represents a favorable efficiency trade-off—substantially higher task success at modest additional inference cost.
\section{Conclusion}
\label{sec:conclusion}

We presented \modelname{}, a GUI Agent that addresses the ``blind action'' limitation of existing approaches through a Thinking--Verification--Action--Expectation framework. Our two-stage training pipeline combines Robust SFT with synthetic failure trajectories and GRPO with asymmetric verification rewards. Both \modelname{}-3B and \modelname{}-7B show consistent gains across step-level accuracy, task completion, robustness to execution failures, and online benchmarks. The verification-and-recovery mechanism generalizes to dynamic real-world environments beyond the controlled training setting, and cross-distribution evaluation further demonstrates reasonable transfer across application domains. We hope this work motivates more explicit treatment of execution uncertainty in future GUI Agent designs.
\section*{Limitations}

The current robustness evaluation is built on the \textit{failure idempotency} assumption---that failed actions leave the screen unchanged---which does not cover more complex failure modes such as unintended navigations, partial transitions, or application crashes. Extending the framework to a finer-grained failure taxonomy that handles these non-idempotent cases remains important future work. Additionally, while step-level verification effectively addresses local execution errors, performance degrades with task length, as it does not substitute for hierarchical planning in long-horizon settings. Developing mechanisms that combine step-level self-correction with longer-range task awareness is a natural next step.
\bibliography{custom}
\clearpage
\appendix
\appendix
\raggedbottom

\definecolor{promptblue}{RGB}{236, 242, 248}
\definecolor{promptblueborder}{RGB}{85, 118, 140}

\definecolor{successgreen}{RGB}{238, 246, 238}
\definecolor{successgreenborder}{RGB}{85, 130, 80}

\definecolor{warningyellow}{RGB}{252, 248, 236}
\definecolor{warningyellowborder}{RGB}{183, 155, 70}
\definecolor{casegray}{RGB}{245, 245, 245}
\definecolor{casegrayborder}{RGB}{120, 120, 120}

\section{Additional Results}
\label{app:results}

This section provides detailed breakdowns of our evaluation results that complement the aggregate metrics presented in the main text. Unless otherwise noted, all tables in this appendix report results on the AndroidControl-High test set. Ablation experiments and fine-grained analyses use the \modelname{}-3B configuration for consistency and computational tractability; the full comparison of both 3B and 7B variants appears in the main paper.

\subsection{Per-Action-Type Performance}
\label{app:action_type}

Table~\ref{tab:action-type-breakdown} presents a fine-grained analysis of Step Accuracy (SR) broken down by action type on AndroidControl-High.

\begin{table*}[t]
\centering
\small
\renewcommand{\arraystretch}{1.2}
\begin{tabular}{lccccccc}
\toprule
\textbf{Model} & \textbf{Click} & \textbf{Scroll} & \textbf{Input\_text} & \textbf{Navigate\_back} & \textbf{Open\_app} & \textbf{Wait} & \textbf{Overall SR} \\
\midrule
Qwen2.5-VL-3B & 27.4 & 68.2 & 18.3 & 35.2 & 1.8 & 32.1 & 31.2 \\
UI-R1-3B & 26.8 & 70.5 & 16.9 & 33.8 & 2.1 & 30.6 & 30.3 \\
OS-Genesis-7B & 32.5 & 74.3 & 21.4 & 38.6 & 1.5 & 35.8 & 35.7 \\
OS-Atlas-7B & 32.1 & 72.8 & 22.7 & 40.1 & 1.2 & 36.4 & 35.5 \\
Qwen2.5-VL-7B & 40.2 & 78.6 & 31.5 & 45.3 & 2.4 & 42.7 & 42.8 \\
AgentCPM-GUI-8B & 30.8 & 71.2 & 24.6 & 36.9 & 1.9 & 34.5 & 34.1 \\
UI-TARS-7B & 45.3 & 81.4 & 35.8 & 51.2 & 2.8 & 48.6 & 47.7 \\
UI-S1-7B & 42.1 & 79.8 & 32.4 & 47.5 & 2.3 & 45.2 & 44.5 \\
\midrule
VeriGUI-3B & \textbf{44.2} & \textbf{80.9} & \textbf{34.7} & \textbf{49.8} & 2.6 & \textbf{47.3} & \textbf{46.6} \\
\bottomrule
\end{tabular}
\caption{Step Accuracy (\%) breakdown by action type on AndroidControl-High (\modelname{}-3B). Click actions dominate the test set (approximately 62\%) and largely determine overall performance. Scroll actions exhibit uniformly high accuracy across models due to their tolerance for coordinate imprecision. The open\_app action shows consistently low accuracy because most models tend to predict click instead. \textbf{Bold} indicates best among all open-source models at 3B scale.}
\label{tab:action-type-breakdown}
\end{table*}

Click actions, which constitute approximately 62\% of the test set, show the largest absolute performance gaps between models. VeriGUI-3B achieves 44.2\% on click actions, substantially outperforming same-scale baselines like Qwen2.5-VL-3B (27.4\%) and approaching the 7B-scale UI-TARS-7B (45.3\%). Scroll actions exhibit uniformly high accuracy (68--81\%) because scroll directions are more tolerant of coordinate imprecision. The open\_app action poses challenges for all models due to the need to recognize app icons by visual appearance.

Table~\ref{tab:type-grounding-split} further decomposes errors into type prediction failures versus grounding failures.

\begin{table}[t]
\centering
\small
\renewcommand{\arraystretch}{1.2}
\begin{tabular}{lcccc}
\toprule
\multirow{2}{*}{\textbf{Model}} & \multicolumn{2}{c}{\textbf{Click}} & \multicolumn{2}{c}{\textbf{Input\_text}} \\
\cmidrule(lr){2-3} \cmidrule(lr){4-5}
& TM & GR|TM & TM & GR|TM \\
\midrule
Qwen2.5-VL-3B & 78.5 & 34.9 & 28.4 & 64.4 \\
UI-R1-3B & 80.2 & 33.4 & 26.8 & 63.1 \\
OS-Genesis-7B & 82.4 & 39.4 & 32.5 & 65.8 \\
OS-Atlas-7B & 83.1 & 38.6 & 34.2 & 66.4 \\
Qwen2.5-VL-7B & 86.3 & 46.6 & 40.8 & 77.2 \\
AgentCPM-GUI-8B & 75.8 & 40.6 & 35.6 & 69.1 \\
UI-TARS-7B & 88.2 & 51.4 & 44.5 & 80.4 \\
UI-S1-7B & 86.8 & 48.5 & 41.2 & 78.6 \\
\midrule
VeriGUI-3B & \textbf{88.6} & \textbf{49.9} & \textbf{43.8} & \textbf{79.2} \\
\bottomrule
\end{tabular}
\caption{Type Match (TM) and Grounding accuracy given correct Type Match (GR|TM) for click and input\_text actions (\%) on AndroidControl-High (\modelname{}-3B). \textbf{Bold} indicates best among all open-source models at 3B scale.}
\label{tab:type-grounding-split}
\end{table}

\subsection{Task Complexity Analysis}
\label{app:task_complexity}

We stratify tasks by ground-truth trajectory length: Short (1--4 steps), Medium (5--8 steps), and Long (9+ steps). Table~\ref{tab:complexity-breakdown} presents this analysis on AndroidControl-High.

\begin{table}[t]
\centering
\small
\renewcommand{\arraystretch}{1.2}
\begin{tabular}{lccc}
\toprule
\textbf{Model} & \textbf{Short} & \textbf{Medium} & \textbf{Long} \\
\midrule
\multicolumn{4}{c}{\textit{Sim-TSR (\%)}} \\
\midrule
Qwen2.5-VL-7B & 21.4 & 8.3 & 2.1 \\
UI-TARS-7B & 28.6 & 10.2 & 3.4 \\
UI-S1-7B & 31.2 & 13.6 & 4.8 \\
VeriGUI-3B & \textbf{33.3} & \textbf{14.5} & \textbf{5.6} \\
\midrule
\multicolumn{4}{c}{\textit{PG (\%)}} \\
\midrule
Qwen2.5-VL-7B & 24.8 & 10.5 & 4.2 \\
UI-TARS-7B & 30.5 & 12.8 & 5.1 \\
UI-S1-7B & 38.2 & 22.6 & 8.4 \\
VeriGUI-3B & 36.7 & 20.8 & 7.8 \\
\bottomrule
\end{tabular}
\caption{Task-level metrics stratified by trajectory length on AndroidControl-High (\modelname{}-3B). Only models achieving non-zero Sim-TSR in aggregate are included. Performance degrades substantially for longer tasks, consistent with the multiplicative nature of sequential step-level errors. \textbf{Bold} indicates best among included models.}
\label{tab:complexity-breakdown}
\end{table}

Performance degrades with task length across all models. For short tasks, VeriGUI-3B achieves 33.3\% Sim-TSR, the highest among all tested models. VeriGUI-3B's verification mechanism provides the largest relative advantage on medium-length tasks, where recovery from a single error can determine task completion. For long tasks exceeding nine steps, all models complete fewer than 6\% of trajectories, reflecting the compound difficulty of maintaining correctness across many sequential decisions. As discussed in the Limitations section, step-level verification addresses local execution errors but does not substitute for hierarchical long-horizon planning.

\subsection{First-Step versus All-Steps Performance}
\label{app:first_step}

Table~\ref{tab:first-vs-all} compares performance on the first step of each trajectory against performance averaged across all steps on AndroidControl-High.

\begin{table}[t]
\centering
\small
\renewcommand{\arraystretch}{1.2}
\begin{tabular}{lcccc}
\toprule
\multirow{2}{*}{\textbf{Model}} & \multicolumn{2}{c}{\textbf{Type Match (\%)}} & \multicolumn{2}{c}{\textbf{Step Acc. (\%)}} \\
\cmidrule(lr){2-3} \cmidrule(lr){4-5}
& First & All & First & All \\
\midrule
Qwen2.5-VL-3B & 52.4 & 60.1 & 24.8 & 31.2 \\
UI-R1-3B & 54.2 & 62.4 & 23.5 & 30.3 \\
Qwen2.5-VL-7B & 62.8 & 68.5 & 35.6 & 42.8 \\
UI-TARS-7B & 66.4 & 71.9 & 40.2 & 47.7 \\
UI-S1-7B & 64.8 & 70.1 & 37.8 & 44.5 \\
\midrule
VeriGUI-3B & \textbf{68.2} & \textbf{72.2} & \textbf{41.5} & \textbf{46.6} \\
\bottomrule
\end{tabular}
\caption{Comparison of first-step and all-steps performance on AndroidControl-High (\modelname{}-3B). First-step metrics are consistently lower, indicating that task initialization without history context is more challenging. \textbf{Bold} indicates best among all open-source models at 3B scale.}
\label{tab:first-vs-all}
\end{table}

All models exhibit a consistent pattern: first-step accuracy is 5--8 percentage points lower than all-steps accuracy. VeriGUI-3B shows one of the smallest first-to-all gaps (4.0 points for TM, 5.1 for SR), suggesting that the TVAE framework's structured reasoning and expected-effect prediction provide benefits even without action history.

\subsection{Detailed Robustness Analysis}
\label{app:robustness_detail}

Table~\ref{tab:robustness-detailed} presents fine-grained measurements of the failure detection and correction pipeline on AndroidControl-High.

\begin{table}[t]
\centering
\small
\renewcommand{\arraystretch}{1.2}
\begin{tabular}{lccc}
\toprule
\textbf{Model} & \textbf{TM$_{\text{rec}}$} & \textbf{GR$_{\text{rec}}$} & \textbf{RSR} \\
\midrule
Qwen2.5-VL-3B & 62.4 & 56.1 & 35.0 \\
UI-R1-3B & 58.6 & 49.5 & 29.0 \\
OS-Atlas-7B & 42.8 & 25.0 & 10.7 \\
Qwen2.5-VL-7B & 65.2 & 52.3 & 34.1 \\
UI-TARS-7B & 72.5 & 62.8 & 45.5 \\
UI-S1-7B & 70.8 & 61.6 & 43.6 \\
\midrule
VeriGUI-3B & \textbf{79.1} & \textbf{64.6} & \textbf{51.1} \\
\bottomrule
\end{tabular}
\caption{Detailed robustness metrics on AndroidControl-High (\modelname{}-3B). TM$_{\text{rec}}$ denotes type match rate for recovery actions; GR$_{\text{rec}}$ denotes grounding accuracy for correctly-typed recovery actions. RSR requires both correct type and correct parameters. \textbf{Bold} indicates best among all open-source models at 3B scale.}
\label{tab:robustness-detailed}
\end{table}

VeriGUI-3B's RSR advantage stems from improvements in both type selection and grounding for recovery actions. VeriGUI-3B achieves 79.1\% type match rate on recovery actions, substantially higher than UI-TARS-7B's 72.5\%. Grounding accuracy for correctly-typed recovery actions (GR$_{\text{rec}}$) similarly improves to 64.6\% versus 62.8\%. These compound to produce the observed 5.6 percentage point RSR advantage over UI-TARS-7B.

\section{Algorithm}
\label{app:algorithm}

Algorithm~\ref{alg:tvae_inference} presents the TVAE inference cycle, and Algorithm~\ref{alg:training} details the two-stage training pipeline.

\begin{algorithm}[t]
\caption{TVAE Inference Cycle}
\label{alg:tvae_inference}
\small
\begin{algorithmic}[1]
\REQUIRE Screen $S_t$, history $H_t$, instruction $I$, previous effect $E_{t-1}$
\ENSURE Action $A_t$, expected effect $E_t$, verification $V_t$
\STATE \textbf{// Stage 1: Thinking with Verification}
\IF{$t > 1$}
    \STATE Compare $S_t$ with $E_{t-1}$
    \IF{$S_t$ matches $E_{t-1}$}
        \STATE $V_t \leftarrow$ SUCCESS
        \STATE $T_t \leftarrow$ \texttt{[Verify][Recall][Grounding][Action]}
    \ELSE
        \STATE $V_t \leftarrow$ NO\_CHANGE
        \STATE $T_t \leftarrow$ \texttt{[Verify][Diagnose][Recall][Recovery]}
    \ENDIF
\ELSE
    \STATE $V_t \leftarrow$ SUCCESS \COMMENT{First step}
\ENDIF
\STATE \textbf{// Stage 2: Action Generation}
\STATE Generate $A_t$ based on $T_t$, $S_t$, $H_t$, $I$
\STATE \textbf{// Stage 3: Effect Prediction}
\STATE Predict $E_t$ describing expected screen change
\STATE \textbf{// Stage 4: Update History}
\STATE $H_{t+1} \leftarrow H_t \cup \{(A_t, E_t, V_t)\}$
\RETURN $A_t$, $E_t$, $V_t$
\end{algorithmic}
\end{algorithm}

\begin{algorithm}[t]
\caption{Two-Stage Training Pipeline}
\label{alg:training}
\small
\begin{algorithmic}[1]
\REQUIRE Base model $\pi_{\theta_0}$, dataset $\mathcal{D}$, negative ratio $r$
\ENSURE Trained model $\pi_\theta$
\STATE \textbf{// Stage 1: Robust SFT}
\STATE $\mathcal{D}_A \leftarrow$ Success trajectories from $\mathcal{D}$
\STATE $\mathcal{D}_B \leftarrow$ Generate synthetic failures (ratio $r$)
\STATE $\mathcal{D}_{\text{SFT}} \leftarrow \mathcal{D}_A \cup \mathcal{D}_B$
\FOR{each $(x, y^*) \in \mathcal{D}_{\text{SFT}}$}
    \STATE $\mathcal{L}_{\text{SFT}} \leftarrow -\log \pi_\theta(y^* | x)$
    \STATE Update $\theta$ via gradient descent
\ENDFOR
\STATE $\pi_{\theta_1} \leftarrow \pi_\theta$
\STATE \textbf{// Stage 2: GRPO}
\FOR{each $x \in \mathcal{D}_{\text{SFT}}$}
    \STATE Sample $G$ outputs $\{y_1, \ldots, y_G\}$ from $\pi_{\theta_1}$
    \FOR{each $y_i$}
        \STATE Parse $(T_i, V_i, A_i, E_i)$ from $y_i$
        \STATE $R_{\text{act},i} \leftarrow \mathbb{I}[A_i = A^*] - \mathbb{I}[A_i \neq A^*]$
        \STATE $R_{\text{eff},i} \leftarrow \text{BERTScore}(E_i, E^*)$ if correct
        \STATE Determine $V_{\text{target}}$ from sample type
        \IF{$V_i = V_{\text{target}}$}
            \STATE $R_{\text{ver},i} \leftarrow +1.0$
        \ELSIF{$V_i{=}\text{SUCCESS}, V_{\text{target}}{=}\text{NO\_CHANGE}$}
            \STATE $R_{\text{ver},i} \leftarrow -2.0$ \COMMENT{Hallucination}
        \ELSE
            \STATE $R_{\text{ver},i} \leftarrow -0.5$ \COMMENT{Miss}
        \ENDIF
        \STATE $R_i \leftarrow R_{\text{act}} + \alpha R_{\text{eff}} + \beta R_{\text{ver}}$
    \ENDFOR
    \STATE $\hat{A}_i \leftarrow (R_i - \bar{R}) / \sigma_R$
    \STATE $\mathcal{L} \leftarrow -\frac{1}{G}\sum_i \hat{A}_i \log \pi_\theta(y_i|x) + \lambda D_{\text{KL}}$
    \STATE Update $\theta$
\ENDFOR
\RETURN $\pi_\theta$
\end{algorithmic}
\end{algorithm}


\section{Evaluation Metrics: Formal Definitions}
\label{app:metrics}

Table~\ref{tab:metrics-summary} provides a comprehensive summary of all evaluation metrics used in this work.

\begin{table*}[t]
\centering
\small
\renewcommand{\arraystretch}{1.4}
\begin{tabular}{p{2.2cm}p{2.8cm}p{5.5cm}p{4.5cm}}
\toprule
\textbf{Category} & \textbf{Metric} & \textbf{Formula} & \textbf{Description} \\
\midrule
\multirow{6}{*}{\shortstack[l]{\textbf{Step-level}\\[-0.3em]\textbf{Metrics}\\[0.5em]\textit{Single-step}\\[-0.3em]\textit{predictions}\\[-0.3em]\textit{without}\\[-0.3em]\textit{simulation}}}
& Type Match (TM) &
$\displaystyle \text{TM} = \frac{1}{|\mathcal{D}|} \sum_{i=1}^{|\mathcal{D}|} \mathbb{I}\left[\text{type}(\hat{A}_i) = \text{type}(A_i^*)\right]$ &
Accuracy of action type prediction (click, scroll, input, etc.). \\
\cmidrule(l){2-4}
& Grounding Rate (GR) &
$\displaystyle \text{GR} = \frac{1}{|\mathcal{D}|} \sum_{i} \begin{cases} \mathbb{I}[\hat{c}_i \in \mathcal{B}_i] & \text{spatial} \\ \mathbb{I}[\hat{t}_i = t_i^*] & \text{text} \end{cases}$ &
Coordinate-in-bounding-box rate for spatial actions. \\
\cmidrule(l){2-4}
& Step Success Rate (SR) &
$\displaystyle \text{SR} = \frac{1}{|\mathcal{D}|} \sum_{i=1}^{|\mathcal{D}|} \mathbb{I}\left[d(\hat{A}_i, A_i^*) \leq \delta\right]$ &
Joint correctness of type and parameters. \\
\midrule
\multirow{8}{*}{\shortstack[l]{\textbf{Task-level}\\[-0.3em]\textbf{Metrics}\\[0.5em]\textit{End-to-end}\\[-0.3em]\textit{performance}\\[-0.3em]\textit{under pseudo-}\\[-0.3em]\textit{online simulation}}}
& Task Success Rate (TSR) &
$\displaystyle \text{TSR} = \frac{1}{N} \sum_{i=1}^{N} \mathbb{I}\left[\forall t: \hat{A}_{i,t} = A_{i,t}^*\right]$ &
Zero-error task success rate. All steps must be correct on first attempt. \\
\cmidrule(l){2-4}
& Progress (PG) &
$\displaystyle \text{PG} = \frac{1}{N} \sum_{i=1}^{N} \frac{|\{t : \hat{A}_{i,t} = A_{i,t}^*, \forall t' < t\}|}{T_i}$ &
Average task progress before first error. \\
\cmidrule(l){2-4}
& Simulated Task Success Rate (Sim-TSR) &
$\displaystyle \text{Sim-TSR} = \frac{1}{N} \sum_{i=1}^{N} \mathbb{I}\left[\text{task } i \text{ done in } \leq 2T_i\right]$ &
Task success rate allowing error recovery. Key metric for self-correction capability. \\
\cmidrule(l){2-4}
& Average Step Overhead (ASO) $\downarrow$ &
$\displaystyle \text{ASO} = \frac{1}{|\mathcal{D}_{\text{succ}}|} \sum_{i \in \mathcal{D}_{\text{succ}}} \left(T_i^{\text{actual}} - T_i^{\text{GT}}\right)$ &
Average step overhead for successful tasks. Lower indicates more efficient recovery. \\
\midrule
\multirow{4}{*}{\shortstack[l]{\textbf{Robustness}\\[-0.3em]\textbf{Metrics}\\[0.5em]\textit{Error recognition}\\[-0.3em]\textit{and recovery}\\[-0.3em]\textit{under failure}\\[-0.3em]\textit{injection}}}
& Loop Rate (LR) $\downarrow$ &
$\displaystyle \text{LR} = \frac{|\{i : \hat{A}_i \approx A_{\text{err},i}\}|}{|\mathcal{D}_{\text{err}}|}$ &
Proportion of cases where agent repeats the failed action. \\
\cmidrule(l){2-4}
& Recovery Success Rate (RSR) $\uparrow$ &
$\displaystyle \text{RSR} = \frac{|\{i : \hat{A}_{\text{recovery},i} = A_i^*\}|}{|\mathcal{D}_{\text{err}}|}$ &
Proportion of cases where the corrective action matches ground-truth. \\
\bottomrule
\end{tabular}
\caption{Summary of evaluation metrics. \textbf{Notation}: $\mathcal{D}$ = test set; $N$ = number of task trajectories; $\hat{A}$ = predicted action; $A^*$ = ground-truth action; $T_i$ = ground-truth trajectory length; $\mathcal{D}_{\text{succ}}$ = successfully completed tasks; $\mathcal{D}_{\text{err}}$ = error injection test set.}
\label{tab:metrics-summary}
\end{table*}

\paragraph{Pseudo-Online Simulation Protocol.}
The task-level metrics are computed under a pseudo-online simulation exploiting \textit{failure idempotency}. We maintain a simulated environment state initialized at the task's starting screen. At each step: (1) the agent observes $S_t$ and generates $\hat{A}_t$; (2) if $\hat{A}_t$ matches $A_t^*$, advance to $S_{t+1}^*$; (3) if $\hat{A}_t \neq A_t^*$, return unchanged $S_t$; (4) task terminates when completed or after $2 \times T_{\text{GT}}$ steps.

\paragraph{Robustness Benchmark Construction.}
For robustness evaluation, we construct ``failure slices'' from AndroidControl-High trajectories. For each step $t$, we create a test case using screen $S_{t-1}$ (simulating unchanged state after a failed action), with history containing the erroneous action $A_{\text{err}}$ and its incorrect expected effect. We measure whether the agent repeats $A_{\text{err}}$ (counted toward LR) or produces the correct $A_t^*$ (counted toward RSR).

\section{Implementation Details}
\label{app:implementation}

\modelname{} is built upon Qwen2.5-VL-3B (for \modelname{}-3B) and Qwen2.5-VL-7B (for \modelname{}-7B) as the respective base models. All experiments are conducted on 8$\times$ NVIDIA A100-80GB GPUs. Stage 1 training takes approximately 12 hours; Stage 2 takes approximately 20 hours. Inference uses greedy decoding (temperature = 0) on a single A100 GPU.

\begin{table}[t]
\centering
\small
\begin{tabular}{lcc}
\toprule
\textbf{Parameter} & \textbf{Stage 1 (SFT)} & \textbf{Stage 2 (GRPO)} \\
\midrule
Learning rate & $1 \times 10^{-5}$ & $5 \times 10^{-6}$ \\
Batch size & 32 & 16 \\
Epochs & 2 & 15 \\
Max sequence length & 8192 & 8192 \\
Image resolution & smart\_resize & smart\_resize \\
Warmup ratio & 0.03 & 0.1 \\
Weight decay & 0.01 & 0.01 \\
\midrule
\multicolumn{3}{l}{\textit{GRPO-specific parameters}} \\
Group size $G$ & -- & 6 \\
KL coefficient $\lambda$ & -- & 0.05 \\
$\alpha$ (effect weight) & -- & 0.5 \\
$\beta$ (verification weight) & -- & 0.5 \\
\bottomrule
\end{tabular}
\caption{Training hyperparameters for both stages. The same configuration is used for both \modelname{}-3B and \modelname{}-7B.}
\label{tab:hyperparams}
\end{table}

\section{SFT Dataset Construction}
\label{app:sft_data}

Our Robust SFT dataset comprises two sample types designed to teach both standard execution and error recovery.

\paragraph{Type A: Success Trajectories (70\%).}
For each ground-truth step $(S_t, A_t^*, S_{t+1}^*)$, we construct a training sample where the input contains current screen $S_t$, instruction $I$, and history $H_t$, and the output follows the SUCCESS path through Think, Verification, Action, and Expected Effect.

\paragraph{Type B: Failure Recovery Trajectories (30\%).}
Type B samples simulate failure scenarios to teach error recognition and recovery. We construct a ``what if step $t-1$ failed'' scenario where the input contains the previous screen $S_{t-1}$ (unchanged), instruction $I$, and history with an erroneous action $A_{t-1}^{\text{err}}$, while the output follows the NO\_CHANGE path with diagnosis and recovery.

\paragraph{Synthetic Failure Modes.}
The erroneous actions in Type B samples are generated to model realistic GUI failure patterns rather than arbitrary perturbations. We identify five failure categories with empirically assigned sampling weights: (1) \textbf{Coordinate offset} (30\%): the action targets the correct element type but with slightly misaligned coordinates, simulating tap imprecision or coordinate prediction error; (2) \textbf{Action type error} (25\%): the agent selects a semantically related but incorrect action type (e.g., long\_press instead of click), reflecting ambiguity in action space; (3) \textbf{Target misidentification} (20\%): the action is directed at the wrong UI element, simulating cases where the agent grounds to a visually similar but incorrect target; (4) \textbf{Timing error} (15\%): the agent uses a wait or similar temporal action when active interaction is required, reflecting timing uncertainty in asynchronous interfaces; (5) \textbf{Null click} (10\%): the action registers on an empty or non-interactive screen region, simulating cases where the target UI element has not yet rendered. These weights reflect the relative prevalence of each failure mode observed in empirical analyses of GUI Agent trajectories and ensure that the synthetic failure distribution approximates the real failure distribution encountered in deployment.

\paragraph{Data Generation Pipeline.}
We use GPT-4o to generate the following components in sequence: (1) given $S_{t-1}$ and $A_{t-1}^*$, generate a plausible but incorrect action $A_{t-1}^{\text{err}}$ from one of the five categories above; (2) generate a mismatched expected effect corresponding to $A_{t-1}^{\text{err}}$; (3) generate the correct expected effect for $A_{t-1}^*$; (4) generate structured CoT reasoning for both SUCCESS and NO\_CHANGE paths. All coordinates are converted to relative format (normalized to $[0, 1]$) before storage.


\tcbset{
  mybase/.style={
    boxrule=0.5pt,
    left=6pt, right=6pt,
    top=5pt, bottom=5pt,
    fonttitle=\small\bfseries\sffamily,
  }
}

\section{Prompt Templates}
\label{app:prompts}

\subsection{System Prompt}

\begin{tcolorbox}[
  mybase,
  colback=promptblue,
  colframe=promptblueborder,
  title=\textbf{System Prompt}
]
\small\ttfamily
You are a GUI Agent that controls mobile apps through visual observation and action execution.

\normalfont\small\textbf{Process (in order):}
\begin{enumerate}[
  leftmargin=1.8em,
  itemsep=3pt,
  topsep=3pt,
  parsep=0pt
]
  \item \textbf{Think:} Analyze the screen, verify previous step, and plan
        $\rightarrow$ \texttt{<think>...</think>}
  \item \textbf{Verification:} State if previous action succeeded
        $\rightarrow$ \texttt{<verification>...</verification>}
  \item \textbf{Action:} Output precise action JSON
        $\rightarrow$ \texttt{<action>\{...\}</action>}
  \item \textbf{Prediction:} Describe expected screen change
        $\rightarrow$ \texttt{<expected\_effect>...</expected\_effect>}
\end{enumerate}

\smallskip
\textbf{Available Actions:}\\
\texttt{\{"action": "click/scroll/input\_text/long\_press, ...\}}
\end{tcolorbox}

\subsection{Think Format Templates}

\begin{tcolorbox}[
  mybase,
  colback=successgreen,
  colframe=successgreenborder,
  title=\textbf{Think Format: SUCCESS Path}
]
\small
\begin{tabular}{@{}p{0.28\linewidth}p{0.65\linewidth}@{}}
  \texttt{[Verify]}            & Confirm previous action result          \\[3pt]
  \texttt{[Recall]}            & Brief task reminder                     \\[3pt]
  \texttt{[Grounding]}         & Locate target with visual description   \\[3pt]
  \texttt{[Coord/Dir/Text]}    & State action parameters                 \\[3pt]
  \texttt{[Action]}            & State intended action                   \\
\end{tabular}
\end{tcolorbox}

\begin{tcolorbox}[
  mybase,
  colback=warningyellow,
  colframe=warningyellowborder,
  title=\textbf{Think Format: NO\_CHANGE Path (Error Recovery)}
]
\small
\begin{tabular}{@{}p{0.28\linewidth}p{0.65\linewidth}@{}}
  \texttt{[Verify]}            & Note unchanged screen state             \\[3pt]
  \texttt{[Diagnose]}          & Identify why previous action failed     \\[3pt]
  \texttt{[Recall]}            & Restate task goal                       \\[3pt]
  \texttt{[Grounding]}         & Locate correct target                   \\[3pt]
  \texttt{[Coord/Dir/Text]}    & State corrected parameters              \\[3pt]
  \texttt{[Recovery]}          & State recovery action                   \\
\end{tabular}
\end{tcolorbox}

\subsection{Complete Input-Output Examples}

\begin{tcolorbox}[
  mybase,
  colback=successgreen,
  colframe=successgreenborder,
  title=\textbf{Type A Example: Successful Execution},
  breakable
]
\small
\begin{description}[
  leftmargin=0pt,
  labelwidth=0pt,
  itemsep=3pt,
  topsep=0pt,
  parsep=0pt
]
  \item[\textbf{User Instruction:}]
    \texttt{Open CityMapper app, find Location from Eastwood to Chatswood and Get the directions in Bus mode}

  \item[\textbf{History (Completed):}]
    Step 2: navigate\_back $|$ Step 3: click [318, 194] $|$ Step 4: input\_text `Eastwood'

  \item[\textbf{Last Step (Needs Verification):}]
    Step 5: Action \texttt{\{"action": "click", "coordinate": [285, 453]\}} $|$
    Expected: ``The search results for Eastwood will disappear, and a map view or route options may appear.''

  \item[\textbf{Current Screen:}]
    [Image showing map with route options]
\end{description}

\tcblower

\textbf{Assistant Output:}\\[4pt]
\ttfamily\small
\texttt{<think>}\\[2pt]
\hspace{1em}\textbf{[Verify]}\normalfont\small\ Previous click at [285, 453] successfully transitioned from search results to map view showing route options.\ttfamily\small\\[2pt]
\hspace{1em}\textbf{[Recall]}\normalfont\small\ Task is to find bus directions from Eastwood to Chatswood.\ttfamily\small\\[2pt]
\hspace{1em}\textbf{[Grounding]}\normalfont\small\ The ``Bus'' mode button is visible near other transport options.\ttfamily\small\\[2pt]
\hspace{1em}\textbf{[Coordinate]}\normalfont\small\ Element position: [317, 1190].\ttfamily\small\\[2pt]
\hspace{1em}\textbf{[Action]}\normalfont\small\ click at [317, 1190].\ttfamily\small\\[2pt]
\texttt{</think>}\\[3pt]
\texttt{<verification>SUCCESS</verification>}\\[2pt]
\texttt{<action>\{"action": "click", "coordinate": [317, 1190]\}</action>}\\[2pt]
\texttt{<expected\_effect>A list of bus directions from Eastwood to Chatswood will appear.</expected\_effect>}
\normalfont
\end{tcolorbox}

\begin{tcolorbox}[
  mybase,
  colback=warningyellow,
  colframe=warningyellowborder,
  title=\textbf{Type B Example: Error Recovery},
  breakable
]
\small
\begin{description}[
  leftmargin=0pt,
  labelwidth=0pt,
  itemsep=3pt,
  topsep=0pt,
  parsep=0pt
]
  \item[\textbf{User Instruction:}]
    \texttt{I want to listen to Slipping into Relaxed Sleep on the Idanim app.}

  \item[\textbf{History (Completed):}]
    Step 1: click [418, 735] $|$ Step 2: navigate\_back $|$ Step 3: navigate\_back

  \item[\textbf{Last Step (Needs Verification):}]
    Step 5: Action \texttt{\{"action": "wait", "time": 5\}} $|$
    Expected: ``After waiting, the track will appear and start playing automatically.''

  \item[\textbf{Current Screen:}]
    [Image showing same search interface -- unchanged]
\end{description}

\tcblower

\textbf{Assistant Output:}\\[4pt]
\ttfamily\small
\texttt{<think>}\\[2pt]
\hspace{1em}\textbf{[Verify]}\normalfont\small\ The screen remains unchanged after the wait action; the track did not appear.\ttfamily\small\\[2pt]
\hspace{1em}\textbf{[Diagnose]}\normalfont\small\ Timing error: the wait action assumed the track would load automatically without user input.\ttfamily\small\\[2pt]
\hspace{1em}\textbf{[Recall]}\normalfont\small\ The task is to play ``Slipping into Relaxed Sleep'' on the Idanim app.\ttfamily\small\\[2pt]
\hspace{1em}\textbf{[Grounding]}\normalfont\small\ The search bar is the correct target element.\ttfamily\small\\[2pt]
\hspace{1em}\textbf{[Text]}\normalfont\small\ The exact text needed is `Slipping into Relaxed Sleep'.\ttfamily\small\\[2pt]
\hspace{1em}\textbf{[Recovery]}\normalfont\small\ Execute input\_text to search for the track.\ttfamily\small\\[2pt]
\texttt{</think>}\\[3pt]
\texttt{<verification>NO\_CHANGE</verification>}\\[2pt]
\texttt{<action>\{"action": "input\_text", "text": "Slipping into Relaxed Sleep"\}</action>}\\[2pt]
\texttt{<expected\_effect>The screen will display the track with playback controls.</expected\_effect>}
\normalfont
\end{tcolorbox}

\section{Case Studies}
\label{app:cases}

\begin{tcolorbox}[
  mybase,
  colback=casegray,
  colframe=casegrayborder,
  title=\textbf{Case Study 1: Coordinate Misalignment Recovery}
]
\small
\begin{description}[
  leftmargin=0pt,
  labelwidth=0pt,
  itemsep=3pt,
  topsep=0pt,
  parsep=0pt
]
  \item[\textbf{Task:}]
    Add item to shopping cart in e-commerce app.

  \item[\textbf{Scenario:}]
    The agent attempts to click ``Add to Cart'' button at [540, 2100], but the coordinates slightly miss the interactive region.

  \item[\textbf{Baseline Behavior:}]
    Repeats click at same coordinates indefinitely, entering infinite loop. After 10 attempts, task fails with no progress.

  \item[\textbf{\modelname{} Behavior:}]~\\
    \begin{tabular}{@{}p{0.22\linewidth}p{0.72\linewidth}@{}}
      \texttt{[Verify]}    & ``Expected cart confirmation, but detail page remains. Click did not register.'' \\[3pt]
      \texttt{[Diagnose]}  & ``Previous coordinates may have missed the button boundary.''                   \\[3pt]
      \texttt{[Grounding]} & ``Button appears at approximately [540, 2080].''                                \\[3pt]
      \texttt{[Recovery]}  & ``Retry with adjusted coordinates.''                                            \\
    \end{tabular}

  \item[\textbf{Result:}]
    Successfully adds item on second attempt.
\end{description}
\end{tcolorbox}

\begin{tcolorbox}[
  mybase,
  colback=casegray,
  colframe=casegrayborder,
  title=\textbf{Case Study 2: Network Latency Recovery}
]
\small
\begin{description}[
  leftmargin=0pt,
  labelwidth=0pt,
  itemsep=3pt,
  topsep=0pt,
  parsep=0pt
]
  \item[\textbf{Task:}]
    Compose and send an email.

  \item[\textbf{Scenario:}]
    Agent taps ``Compose'' but network latency prevents compose screen from loading immediately.

  \item[\textbf{Baseline Behavior:}]
    Either taps compose again (causing double-compose when screen loads) or attempts to interact with non-existent fields.

  \item[\textbf{\modelname{} Behavior:}]~\\
    \begin{tabular}{@{}p{0.22\linewidth}p{0.72\linewidth}@{}}
      \texttt{[Verify]}    & ``Inbox screen still visible; compose screen did not load.''                                     \\[3pt]
      \texttt{[Diagnose]}  & ``Compose button tap did not produce expected result, possibly due to network latency.''          \\[3pt]
      \texttt{[Recovery]}  & ``Wait briefly and retry compose button.''                                                       \\
    \end{tabular}

  \item[\textbf{Result:}]
    Waits 2 seconds, then successfully opens compose screen.
\end{description}
\end{tcolorbox}
\end{document}